%% file: main.tex
\documentclass[11pt]{article}

\usepackage[preprint]{acl}
\usepackage{times}
\usepackage{latexsym}
\usepackage[T1]{fontenc}
\usepackage[utf8]{inputenc}
\usepackage{fix-cm}
\usepackage{anyfontsize}
\usepackage{microtype}
\usepackage{inconsolata}
\usepackage{graphicx}
\usepackage{booktabs}
\usepackage{amsmath}
\usepackage{amssymb}
\usepackage{tikz}
\usepackage{float}
\usepackage{placeins}
\usepackage{stfloats}

\title{DAMI: Dual-Axis Memory Initialization for Enhanced Long-Context Retrieval in Gated Linear Attention}

\author{
  Zhangchi Feng\textsuperscript{1,2}\quad
  Harry Jiang\textsuperscript{2}\quad
  Yangyifei Luo\textsuperscript{1} \\
  \bfseries Ray Zhang\textsuperscript{2}\quad
  Richong Zhang\textsuperscript{2,*} \\
  \normalfont\textsuperscript{1}SpeedAI \\
  \textsuperscript{2}CCSE, School of Computer Science and Engineering, \\
  Beihang University \\
  \texttt{muller@speedai.com}\quad\texttt{2409281700@qq.com} \\
  \texttt{luoyangyifei@speedai.com}\quad\texttt{2227613285@qq.com} \\
  \normalfont\textsuperscript{*}Corresponding author: \texttt{zhangrc@act.buaa.edu.cn}
}
\begin{document}
\maketitle

\begin{abstract}
Linear attention scales linearly with sequence length, but its finite-state
memory can limit long-context retrieval. In gated linear recurrences, memory
retention depends on state decay, which is shaped jointly by gate parameters
\texttt{A\_log} and \texttt{dt\_bias} and by input activations. In controlled
Gated DeltaNet (GDN) experiments, both parameters remain highly correlated
with their initial values after training, highlighting the importance of initialization. Therefore, we introduce Dual-Axis Memory
Initialization (DAMI), which jointly configures these parameters before
training. DAMI assigns geometric timescales across heads for varied retrieval
distances and a low--high--low input-sensitivity profile across layers. A
small activation sample calibrates the GDN-family implementation to match
both behavioral targets without adding trainable parameters. Across GDN, KDA, GDN2, Mamba-2, and RWKV-7, DAMI
improves average RULER-style retrieval accuracy by 5.29--16.71 percentage
points over the corresponding random-initialization baselines. Performance on
language modeling, commonsense reasoning, and 2K real-world retrieval remains
broadly comparable to those baselines.\footnote{Our code is available at
\url{https://speedai.com/research/dami}.}
\end{abstract}

\input{sections/01-introduction}
\input{sections/02-related-work}
\input{sections/03-method}
\input{sections/04-experiments}
\input{sections/05-conclusion}
\input{sections/06-limitations}

\FloatBarrier
\bibliography{references}

\clearpage
\appendix
\input{sections/appendix}

\end{document}

%% file: sections/01-introduction.tex
\section{Introduction}
\label{sec:introduction}

\begin{figure}[!t]
\centering
\includegraphics[width=\columnwidth]{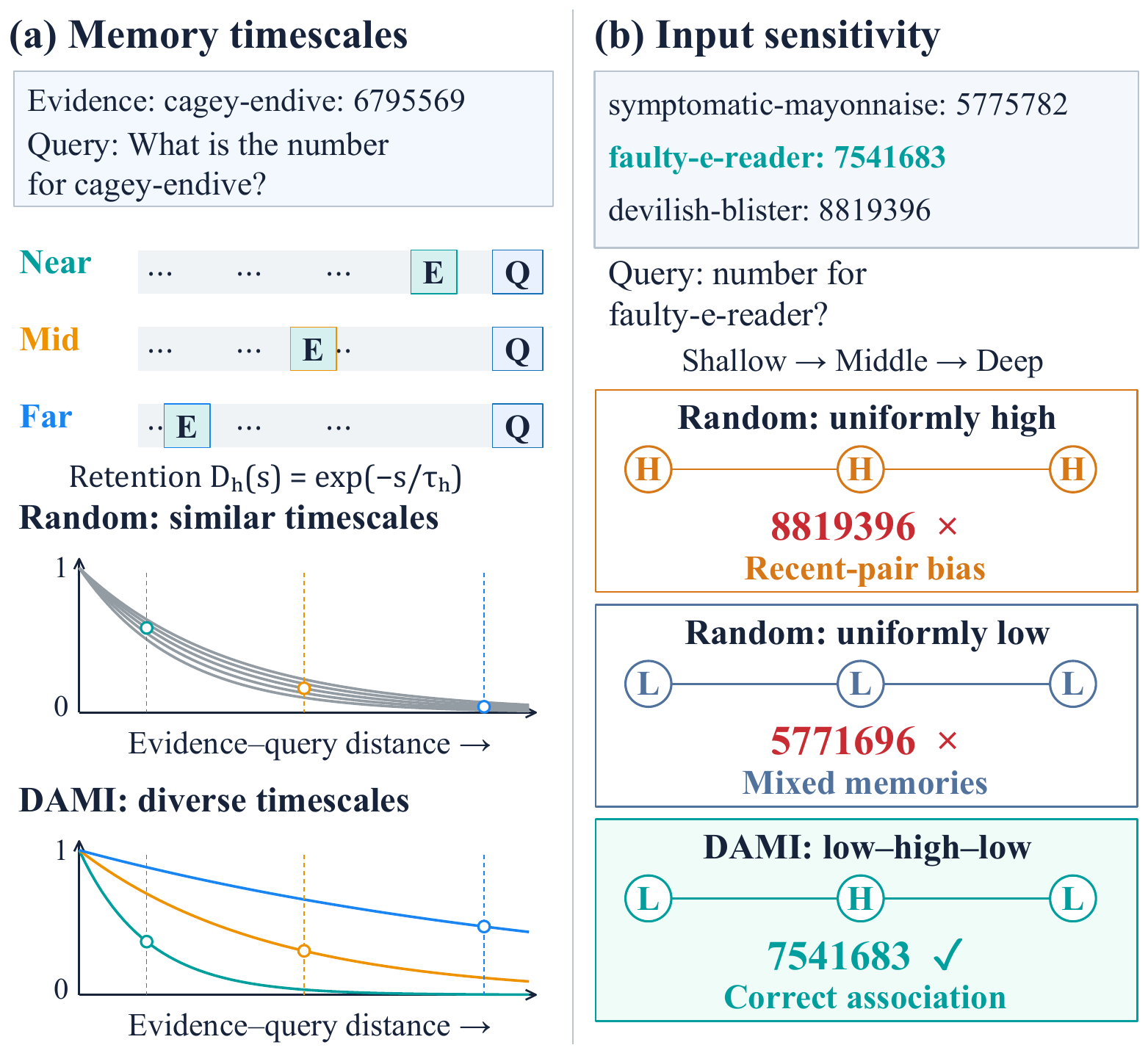}
\caption{Dual-axis memory initialization.
(a) Diverse head-wise timescales broaden memory coverage across evidence--query
distances; curves illustrate normalized exponential decay.
(b) Schematic uniform-high and uniform-low Random regimes illustrate recent-pair
bias and mixed memories, contrasted with DAMI's low--high--low sensitivity profile.
Text excerpts are from RULER; placements and outcomes are schematic.}
\label{fig:dami-teaser}
\end{figure}

Long-context retrieval requires retaining distant information despite
intervening distractors. Linear attention and gated recurrences replace
softmax attention's quadratic computation and growing cache with fixed-size
states \citep{vaswani2017attention,katharopoulos2020transformers}.
Retrieval remains challenging even as recurrent architectures advance
\citep{arora2024zoology,arora2024based,yang2025gdn,kimi2025linear,hatamizadeh2026gdn2}.
We investigate decay-gate initialization as a means of improving memory
without changing state-update operators.

In controlled Gated DeltaNet (GDN) runs, the learnable memory parameters
\texttt{A\_log} and \texttt{dt\_bias} remain highly correlated with their
initial values (Section~\ref{sec:initialization-role}), motivating structured
initialization. PoST improves long-range memory through geometric decay
spectra and position-adaptive scaling \citep{cao2026optimal}.
Realized decay, however, depends jointly on the decay multiplier
\texttt{A\_log}, the Softplus operating region shifted by \texttt{dt\_bias},
and input activations. Achieving intended memory behavior therefore requires
coordinating both parameters.

We propose Dual-Axis Memory Initialization (DAMI): geometric target
timescales across heads and a low--high--low input-sensitivity profile
across layers. Approximate power-law dependencies in text and their
exponential-mixture representation motivate multiscale coverage
\citep{altmann2012longrange,lin2017critical,beylkin2010revisited}, while
different representation stages motivate depth-dependent responses.
Figure~\ref{fig:dami-teaser} illustrates these complementary requirements.
DAMI jointly calibrates both targets from actual activations before
training. Its GDN-family implementation adjusts native memory parameters;
Mamba-2 and RWKV-7 use architecture-specific extensions
(Section~\ref{sec:architecture-extensions}).

Across approximately 100M-parameter GDN, KDA, GDN2, Mamba-2, and RWKV-7
models trained on 5B tokens, DAMI improves average RULER-style NIAH accuracy
\citep{hsieh2024ruler} by 5.29--16.71 percentage points over Random and
outperforms PoST-style spectral initialization. Language modeling,
commonsense reasoning, and real-world retrieval remain broadly comparable.

Our main contributions are:
\begin{itemize}
\item We identify persistent correlations between initial and trained GDN
memory parameters, motivating structured initialization.
\item We introduce DAMI, jointly configuring head-wise timescales and
layer-wise sensitivities, with extensions to Mamba-2 and RWKV-7.
\item Five-architecture experiments demonstrate improved retrieval with
broadly comparable general capabilities; GDN ablations establish the
benefit of the joint configuration.
\end{itemize}

%% file: sections/02-related-work.tex
\section{Related Work}
\label{sec:related-work}

\paragraph{Gated Recurrences and Long-Context Retrieval.}
Linear attention compresses history into a fixed-size state
\citep{katharopoulos2020transformers}. GLA, DeltaNet, and the GDN family improve
recurrence through gating and the delta rule
\citep{yang2024gla,yang2024deltanet,yang2025gdn,kimi2025linear,hatamizadeh2026gdn2};
Mamba develops selective state spaces and state space duality
\citep{gu2024mamba,dao2024mamba2}. 
Zoology and RULER distinguish computational efficiency from effective
recall through retrieval evaluations
\citep{arora2024zoology,hsieh2024ruler}. DeciMamba and LongMamba extend receptive
fields through post-training filtering
\citep{benkish2024decimamba,ye2025longmamba}. 
DAMI operates at initialization, leaving the recurrence and inference
procedure unchanged.

\paragraph{Decay Spectra and Power-Law Dependencies.}
Multiscale memory has also been explored in dynamic graphs, where hierarchical
memories combine continuous-time updates with multiple discrete-time granularities
to capture temporal dynamics \citep{zhang2022hierarchical}.
HiPPO, S4, and S4D connect long-range memory to structured dynamics and spectra
\citep{gu2020hippo,gu2022s4,gu2022s4d}. Approximate power-law dependencies in text
\citep{altmann2012longrange,lin2017critical} motivate exponential mixtures with
controlled approximation error
\citep{beylkin2005exponential,beylkin2010revisited}. PoST applies geometric
spectra and position-adaptive scaling to linear recurrences
\citep{cao2026optimal}. 
Building on this multiscale motivation, DAMI jointly realizes target
timescales and input responses in GDN gates.

\paragraph{Memory Initialization.}
Chrono initialization uses anticipated temporal ranges, while LRU combines
stable parameterization, eigenvalue initialization, and normalization
\citep{tallec2018chrono,orvieto2023lru}. LSUV calibrates activation variance
sequentially using real samples \citep{mishkin2016lsuv}. DAMI uses this
layer-sequential approach to calibrate head-wise timescales and layer-wise
Softplus sensitivities:
it jointly determines \texttt{A\_log} and \texttt{dt\_bias} before training.

%% file: sections/03-method.tex
\section{Method}
\label{sec:method}

\begin{figure*}[!t]
\centering
\includegraphics[width=\textwidth]{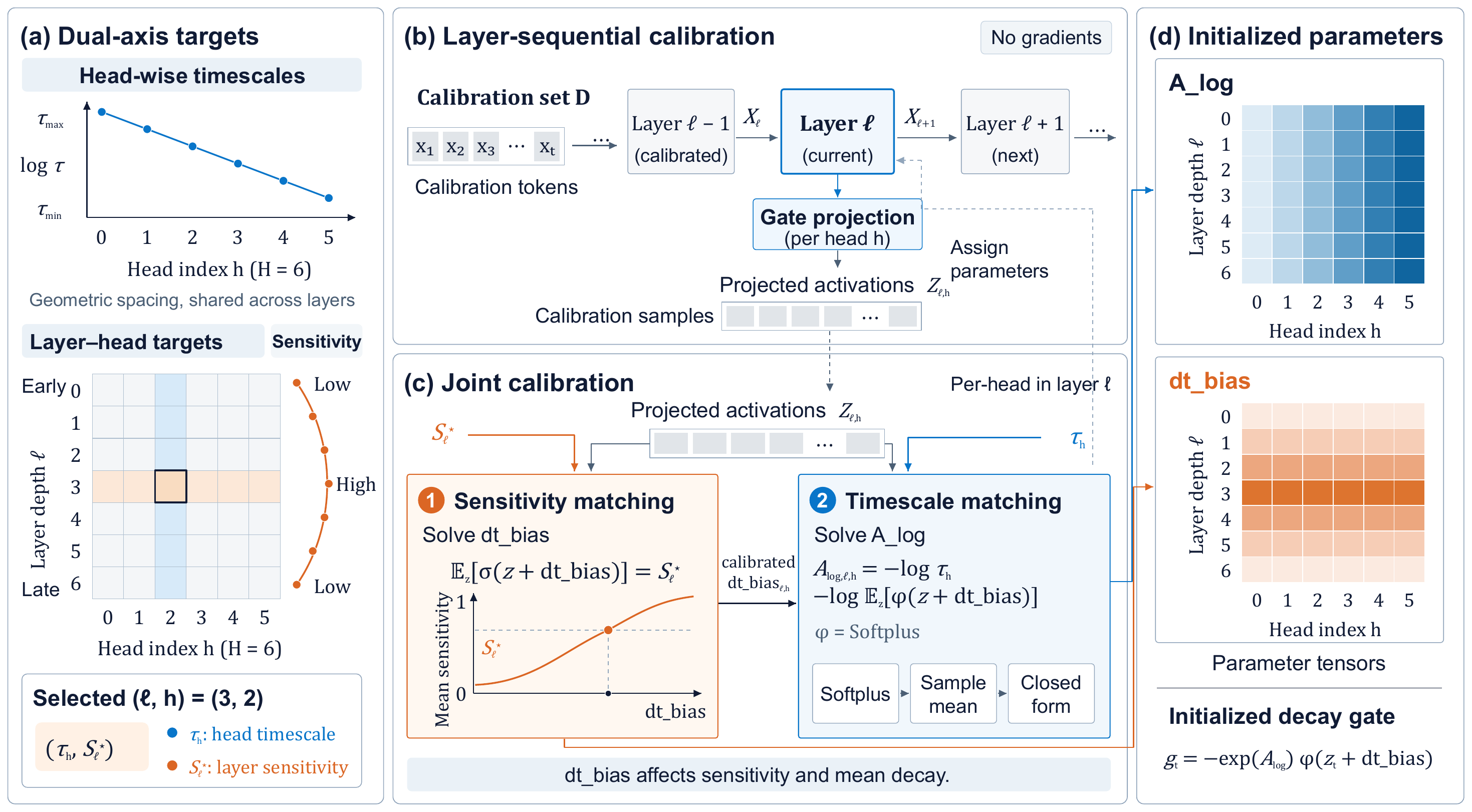}
\caption{Overview of DAMI initialization. (a) Head-wise timescales and
layer-wise sensitivities define dual-axis targets. (b) Calibration proceeds
layer by layer using projected activations without gradients. (c) Each head's
\texttt{dt\_bias} is solved before \texttt{A\_log} using calibration-set
statistics. (d) Schematic parameter layouts illustrate the head and layer axes.}
\label{fig:dami-overview}
\end{figure*}

We formulate head-wise timescales and layer-wise Softplus sensitivities for
finite-state recurrences \citep{katharopoulos2020transformers,dao2024mamba2},
motivated by persistent initialization structure.
Figure~\ref{fig:dami-overview} summarizes their joint calibration in DAMI.

\subsection{Preliminary}
\label{sec:recurrent-memory}

Let \(S_{t,\ell,h}\in\mathbb{R}^{d_k\times d_v}\) denote the recurrent state at
position \(t\) for head \(h\) in layer \(\ell\). In Gated DeltaNet (GDN)
\citep{yang2025gdn}, the update can be written as
\begin{equation}
\begin{aligned}
\widetilde S_{t-1}&=\exp(g_t)S_{t-1}, \\
S_t&=\widetilde S_{t-1}
    +\beta_t k_t\bigl(v_t^{\top}-k_t^{\top}\widetilde S_{t-1}\bigr).
\end{aligned}
\label{eq:gdn-update}
\end{equation}
We omit layer and head indices for brevity. Here \(k_t,v_t\) are key and value,
\(\beta_t\) is a scalar update coefficient, and \(\widetilde S_{t-1}\) is the
decayed state. The query reads \(o_t=S_t^\top q_t\).
The explicit retention factor is \(\exp(g_t)\), with
\begin{equation}
\begin{aligned}
g_{t,\ell,h}&=-\exp(A_{\log,\ell,h}) \\
&\quad\cdot\operatorname{softplus}\!\left(z_{t,\ell,h}+\mathrm{dt\_bias}_{\ell,h}\right), \\
z_{t,\ell,h}&=\operatorname{proj}(x_t)_{\ell,h}.
\end{aligned}
\label{eq:gate}
\end{equation}
The two parameters to initialize are \texttt{A\_log} and \texttt{dt\_bias}.
Both are independently parameterized per head in GDN: every head in every layer
has its own scalar \texttt{A\_log} and \texttt{dt\_bias}. Together, they determine
historical forgetting and the gate's response to projected activations.

KDA \citep{kimi2025linear} and GDN2 \citep{hatamizadeh2026gdn2} retain one
\texttt{A\_log} scalar per head but use key-channel
\(\mathrm{dt\_bias}_{\ell,h,k}\), rather than GDN's head-level scalar.
Writing \(\widetilde S_{t-1}=D_tS_{t-1}\), GDN uses \(D_t=e^{g_t}I\);
KDA/GDN2 use \(D_t=\operatorname{Diag}(e^{g_t})\).
Mamba-2 \citep{dao2024mamba2} also uses head-wise Softplus decay, whereas
RWKV-7 \citep{peng2025rwkv7} uses a sigmoid decay gate. Table~\ref{tab:gdn-family}
summarizes their distinct operators and the parameterizations evaluated here.

In the GDN family, native initialization samples \(\exp(A_{\log})\) uniformly and takes its
logarithm to obtain \texttt{A\_log}, using \([0,16)\) for GDN and \([1,16)\)
for the default KDA implementation and GDN2. For \texttt{dt\_bias}, values are
sampled log-uniformly from \([10^{-3},10^{-1}]\) and transformed by inverse
Softplus. The same sampling rules are used across layers, independently of
input statistics.\footnote{See the
\href{https://github.com/NVlabs/GatedDeltaNet/blob/main/lit_gpt/gated_delta_net.py}{official GDN implementation},
\href{https://github.com/fla-org/flash-linear-attention/blob/main/fla/layers/kda.py}{default FLA implementation of KDA},
and \href{https://github.com/NVlabs/GatedDeltaNet-2/blob/main/lit_gpt/gdn2.py}{official GDN2 implementation}.}

We focus our analysis, initialization design, and ablations on GDN, and further
evaluate DAMI's cross-architecture applicability on KDA, GDN2, Mamba-2,
and RWKV-7. The latter two adaptations are described in
Section~\ref{sec:architecture-extensions}.

\begin{table*}[!t]
\centering
\fontsize{6.5}{8}\selectfont
\setlength{\tabcolsep}{1pt}
\setlength{\thinmuskip}{2mu}
\setlength{\medmuskip}{2mu}
\setlength{\thickmuskip}{3mu}
\renewcommand{\arraystretch}{1.18}
\begin{tabular*}{\textwidth}{@{\extracolsep{\fill}}l c c c c l@{}}
\toprule
Model & Gate \(g_t\) & \texttt{A\_log} & \texttt{dt\_bias} & Decayed state & State update \\
\midrule
GDN & \(-e^{A_{\log}}\phi(z_t+\mathrm{dt\_bias})\)
& Head-level & Head-level
& \(\widetilde S_{t-1}=e^{g_t}S_{t-1}\)
& \(S_t=\widetilde S_{t-1}+\beta_t k_t(v_t^{\top}-k_t^{\top}\widetilde S_{t-1})\) \\
KDA & \(-e^{A_{\log}}\phi(z_{t,k}+\mathrm{dt\_bias}_k)\)
& Head-level & Key-channel
& \(\widetilde S_{t-1}=\operatorname{Diag}(e^{g_t})S_{t-1}\)
& \(S_t=\widetilde S_{t-1}+\beta_t k_t(v_t^{\top}-k_t^{\top}\widetilde S_{t-1})\) \\
GDN2 & \(-e^{A_{\log}}\phi(z_{t,k}+\mathrm{dt\_bias}_k)\)
& Head-level & Key-channel
& \(\widetilde S_{t-1}=\operatorname{Diag}(e^{g_t})S_{t-1}\)
& \(S_t=\widetilde S_{t-1}+k_t[(w_t\odot v_t)^{\top}-(b_t\odot k_t)^{\top}\widetilde S_{t-1}]\) \\
Mamba-2 & \(-e^{A_{\log}}\phi(z_t+\mathrm{dt\_bias})\)
& Head-level & Head-level
& \(\widetilde S_{t-1}=e^{g_t}S_{t-1}\)
& \(S_t=\widetilde S_{t-1}+\Delta_t B_t x_t^{\top}\) \\
RWKV-7 & \(-e^{A_{\log}}\sigma(z_{t,k}+\mathrm{dt\_bias}_k)\)
& Head-level (frozen) & Key-channel
& \(\widetilde S_{t-1}=\operatorname{Diag}(e^{g_t})S_{t-1}\)
& \(S_t=\widetilde S_{t-1}+\tilde k_t v_t^{\top}-p_t(\hat\kappa_t^{\top}S_{t-1})\) \\
\bottomrule
\end{tabular*}
\caption{Gate parameterizations and state updates.
\(\phi=\operatorname{softplus}\), \(\sigma\) is sigmoid, and \(\odot\) denotes
element-wise multiplication. \(\beta_t\) is scalar; \(b_t,w_t\) are independent
channel-wise coefficients. For Mamba-2,
\(\Delta_t=\phi(z_t+\mathrm{dt\_bias})\), with input vectors \(B_t,x_t\).
For RWKV-7, \(p_t=a_t\odot\hat\kappa_t\), with normalized key
\(\hat\kappa_t\) and channel-wise coefficient \(a_t\).}
\label{tab:gdn-family}
\end{table*}

\subsection{Persistence of Initialization Structure}
\label{sec:initialization-role}

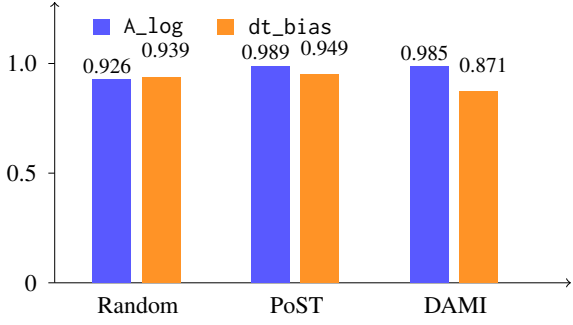
\begin{figure}[!htb]
\centering
\begin{tikzpicture}[x=1.05cm,y=2.9cm,font=\fontsize{9}{10.5}\selectfont]
  \draw[->,line width=0.35pt] (0,0) -- (6.5,0);
  \draw[->,line width=0.35pt] (0,0) -- (0,1.28);
  \foreach \y/\ylabel in {0/0,0.5/0.5,1.0/1.0}
    \draw[line width=0.25pt] (-0.08,\y) -- (0,\y) node[left=3pt] {\ylabel};
  \foreach \x/\xlabel in {1.05/{Random},3.05/{PoST},5.05/{DAMI}}
    \node[align=center] at (\x,-0.10) {\xlabel};
  \fill[blue!65] (0.48,0) rectangle (0.96,0.926);
  \fill[orange!85] (1.10,0) rectangle (1.58,0.939);
  \fill[blue!65] (2.48,0) rectangle (2.96,0.989);
  \fill[orange!85] (3.10,0) rectangle (3.58,0.949);
  \fill[blue!65] (4.48,0) rectangle (4.96,0.985);
  \fill[orange!85] (5.10,0) rectangle (5.58,0.871);
  \foreach \x/\v/\offset in {0.66/0.926/0.025,1.40/0.939/0.085,2.66/0.989/0.025,3.40/0.949/0.085,4.66/0.985/0.025,5.40/0.871/0.085}
    \node[anchor=south,font=\fontsize{8}{9}\selectfont,inner sep=0pt] at (\x,{\v+\offset}) {\v};
  \fill[blue!65] (0.48,1.13) rectangle (0.69,1.20);
  \node[anchor=west,font=\fontsize{9}{10.5}\selectfont] at (0.74,1.165) {\texttt{A\_log}};
  \fill[orange!85] (2.05,1.13) rectangle (2.26,1.20);
  \node[anchor=west,font=\fontsize{9}{10.5}\selectfont] at (2.31,1.165) {\texttt{dt\_bias}};
\end{tikzpicture}
\caption{Initial--trained Pearson correlations of GDN memory parameters.}
\label{fig:initialization-correlation}
\end{figure}

Following long-range initialization studies \citep{gu2022s4d,orvieto2023lru},
we compare initial and trained GDN memory parameters from our main runs.
Pearson correlations for \texttt{A\_log} and \texttt{dt\_bias} are
0.926/0.939 for \textsc{Random}, 0.989/0.949 for \textsc{PoST}, and
0.985/0.871 for DAMI (Figure~\ref{fig:initialization-correlation}).
These high correlations indicate that trained memory parameters retain
statistical structure associated with initialization.

Our GDN retrieval comparisons fix model size, data, optimizer, and training
budget; DAMI copies the corresponding \textsc{Random} model's non-memory
parameters. The joint dependence of decay on \texttt{A\_log},
\texttt{dt\_bias}, and activations (Equation~\eqref{eq:gate}) motivates
coupled behavioral targets.

\subsection{Head-wise Timescale Allocation}
\label{sec:alog-timescale}
\label{sec:gdn-decay-derivation}
\label{sec:powerlaw-basis}

GDN applies \(\exp(g_{t,\ell,h})\) at each step. Taking the initial memory
position as local time zero, let \(s\) count subsequent tokens.
The proportion \(D_{\ell,h}(s)\) retained along the explicit decay path is
\begin{equation}
\begin{aligned}
D_{\ell,h}(s)&=\prod_{t=1}^{s}\exp(g_{t,\ell,h})\\
&=\exp\!\left[-\exp(A_{\log,\ell,h})L_{\ell,h}(s)\right],\\
L_{\ell,h}(s)&=\sum_{t=1}^{s}
\operatorname{softplus}(z_{t,\ell,h}+\mathrm{dt\_bias}_{\ell,h}).
\end{aligned}
\label{eq:retention}
\end{equation}
During inference, \(\exp(A_{\log,\ell,h})\) is fixed, while each new token adds
a Softplus term to \(L_{\ell,h}(s)\). The multiplier and the input response
therefore jointly control this envelope, which tracks multiplicative decay
with interference from subsequent inputs set aside.

The retention factor \(D_{\ell,h}(s)\) quantifies the fraction of the initial
contribution remaining after \(s\) tokens. We define the \emph{e-folding time}
\(\tau_{\ell,h}\) as the token interval required for this proportion to reach
\(1/e\). When one-step decay is approximately constant, cumulative retention
has the standard exponential form
\begin{equation}
D_{\ell,h}(s)\approx\exp(-s/\tau_{\ell,h}).
\label{eq:timescale}
\end{equation}
Larger \(\tau_{\ell,h}\) implies slower decay; at \(s=\tau_{\ell,h}\), the
envelope is approximately \(1/e\).

Natural-language dependencies span multiple distances:
\citet{altmann2012longrange} connect long-range correlations to hierarchical
structure, and \citet{lin2017critical} observe approximate power-law decay of
character mutual information in English and French, motivating multiscale
memory coverage.

Under approximately constant one-step decay, a head supplies a single
exponential timescale. Cross-scale coverage instead motivates a mixture,
using the Laplace representation and finite-range approximation
\citep{beylkin2005exponential,beylkin2010revisited}:
\begin{equation}
\begin{aligned}
s^{-\beta}&=\frac{1}{\Gamma(\beta)}
\int_{0}^{\infty}\lambda^{\beta-1}\exp(-\lambda s)\,\mathrm{d}\lambda \\
&\approx\sum_{h=0}^{H-1}w_h\exp(-\lambda_h s), \quad s>0,\;\beta>0.
\end{aligned}
\label{eq:powerlaw-mixture}
\end{equation}
Here \(\lambda_h\) is an exponential component's decay rate and \(w_h\) its
weight. GDN's multihead output provides a structural counterpart to this mixture.


Each layer concatenates head outputs and applies an output projection
\citep{vaswani2017attention}. Let \(\tilde{o}_{t,\ell,h}\) denote the vector
from head \(h\), including preceding normalization or gating. Then

Partitioning \(W_{O,\ell}=[\,W_{O,\ell,0}\ \cdots\ W_{O,\ell,H-1}\,]\) by head gives
\begin{equation}
\begin{aligned}
y_{t,\ell}&=W_{O,\ell}\operatorname{Concat}_{h=0}^{H-1}
(\tilde{o}_{t,\ell,h})\\
&=\sum_{h=0}^{H-1}W_{O,\ell,h}\tilde{o}_{t,\ell,h}.
\end{aligned}
\label{eq:output-blocks}
\end{equation}
Equation~\eqref{eq:output-blocks} combines head components through projection
blocks, resembling Equation~\eqref{eq:powerlaw-mixture}.
Separating content-dependent factors in an envelope approximation gives
effective coefficients \(c_{\ell,h}\) and temporal structure
\begin{equation}
K_{\ell}(s)=\sum_{h=0}^{H-1}c_{\ell,h}\exp(-s/\tau_{\ell,h}).
\label{eq:head-mixture}
\end{equation}
The effective coefficients \(c_{\ell,h}\) summarize content-dependent head
contributions and their linear combination by the output projection at the
envelope level.

Theorem~4.7 of PoST \citep{cao2026optimal} supports geometric spacing through
an exponential error rate for finite-range power-law approximation
(Appendix~\ref{app:spectral}). We therefore use
\begin{equation}
\begin{aligned}
\lambda_h&=\lambda_{\min}
\left(\frac{\lambda_{\max}}{\lambda_{\min}}\right)^{\frac{h}{H-1}}, \\
\tau_h&=\lambda_h^{-1}, \qquad h=0,\ldots,H-1,
\end{aligned}
\label{eq:geometric-spectrum}
\end{equation}
Equation~\eqref{eq:retention} links these target timescales to \texttt{A\_log}
through activations and \texttt{dt\_bias}.

The input-response target is defined below, and both parameters are jointly
calibrated in Section~\ref{sec:dami}.

\subsection{Layer-wise Input Sensitivity}
\label{sec:dtbias-response}

Timescales describe decay speed, but \texttt{dt\_bias} also changes the
Softplus factor's response to activations. Its mean local sensitivity is
defined as
\begin{equation}
\begin{aligned}
\mathcal{S}_{\ell,h}(\mathrm{dt\_bias})
&:=\mathbb{E}_{z}\left[\sigma(z+\mathrm{dt\_bias})\right].
\end{aligned}
\label{eq:sensitivity}
\end{equation}
Since Softplus has derivative \(\sigma(z+\mathrm{dt\_bias})\), lower
\(\mathcal{S}\) means a weaker response of this factor.

\begin{figure}[!htb]
\centering
\includegraphics[width=\columnwidth]{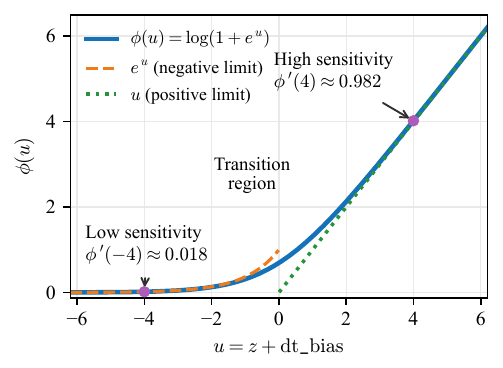}
\caption{Effect of \texttt{dt\_bias} on local Softplus sensitivity. Dashed and
dotted curves show the asymptotic approximations;
\(\phi'(-4)\approx0.018\) and \(\phi'(4)\approx0.982\).}
\label{fig:softplus-sensitivity}
\end{figure}

Figure~\ref{fig:softplus-sensitivity} shows how bias moves the operating point
along Softplus: its slope approaches zero in the negative regime and one in
the positive regime, changing the factor's response to a fixed activation.

Native GDN applies the same sampling rule across layers. Yet each layer's
activation distribution is shaped by preceding projections, convolutions,
and recurrent updates. Together with \texttt{dt\_bias}, these distributions
determine sensitivity in Equation~\eqref{eq:sensitivity}, motivating
depth-dependent response targets.

We adopt a low--high--low profile to balance token-level stability,
content-dependent adaptation, and long-term retention: early layers receive
low sensitivity, middle layers high sensitivity, and late layers low
sensitivity. Section~\ref{sec:ablations} evaluates this allocation.

We assign a target sensitivity \(\mathcal{S}^{\star}_{\ell}\) to each layer,
complementing the head-wise timescale spectrum in
Section~\ref{sec:alog-timescale}.

\subsection{Joint Data-dependent Initialization}
\label{sec:dami}

As illustrated in Figure~\ref{fig:dami-overview}(b)--(c), layers share head-wise
timescales \(\{\tau_h\}_{h=0}^{H-1}\), while heads within each layer share a
sensitivity target \(\mathcal{S}^{\star}_{\ell}\).
Following the layer-sequential, data-driven approach of LSUV
(Layer-Sequential Unit-Variance), introduced in \emph{All you need is a good init}
\citep{mishkin2016lsuv}, DAMI calibrates sensitivity and timescale from shallow
to deep layers.

For each layer, we collect projected activations \(Z_{\ell,h}\) without
gradients and first solve by bisection
\begin{equation}
\frac{1}{|Z_{\ell,h}|}\sum_{z\in Z_{\ell,h}}
\sigma(z+\mathrm{dt\_bias}_{\ell,h})=\mathcal{S}^{\star}_{\ell}.
\label{eq:solve-dt}
\end{equation}

We then match the target timescale using the mean Softplus factor:
\begin{equation}
\begin{aligned}
A_{\log,\ell,h}&=-\log\tau_h\\
&\quad-\log\mathbb{E}_{z\sim Z_{\ell,h}}\!\left[
\phi(z+\mathrm{dt\_bias}_{\ell,h})\right].
\end{aligned}
\label{eq:solve-alog}
\end{equation}

Here \(\phi=\operatorname{softplus}\) and
\(\mathbb{E}_{z\sim Z_{\ell,h}}\) denotes the sample mean.
Equation~\eqref{eq:solve-alog} compensates for the bias's effect on mean decay,
coordinating the two targets; Appendix~\ref{app:calibration} gives the derivation.
The calibrated output feeds the next layer. The output projection retains
its native initialization and learns to combine heads during training.

\paragraph{Architecture-specific extensions.}
\label{sec:architecture-extensions}
KDA/GDN2 retain channel-wise \texttt{dt\_bias}, matching within-head means
with small centered perturbations. Mamba-2 retains geometric head-wise
spectra and scales static Softplus biases along depth to transfer the
low--high--low profile. RWKV-7 instead schedules the static biases' mean
sigmoid response and matches head-wise timescales; all three conditions use
the same 160 frozen head--layer scales without adding trainable parameters.
All adaptations preserve their state-update operators.
Appendices~\ref{app:extensions} and~\ref{app:settings} give details.

%% file: sections/04-experiments.tex
\section{Experiments}
\label{sec:experiments}

We evaluate retrieval and general capabilities, ablate both initialization
axes, and examine retrieval across distances.

\subsection{Experimental Setup}
\label{sec:experimental-setup}

\paragraph{Models and training.}
We evaluate GDN \citep{yang2025gdn}, KDA \citep{kimi2025linear}, GDN2
\citep{hatamizadeh2026gdn2}, Mamba-2 \citep{dao2024mamba2}, and RWKV-7
\citep{peng2025rwkv7} at approximately 100M parameters, each trained on 5B
tokens with a maximum context of 4,096 tokens.
Within each architecture, comparisons share parameterization and training
recipe; GDN-family and Mamba-2 variants preserve Random's non-decay
initialization. RWKV-7 uses the common frozen-scale parameterization in
Section~\ref{sec:architecture-extensions}; its selected Random/PoST
checkpoints use seed 42 and DAMI uses seed 43.
Architecture dimensions are in Appendix~\ref{app:settings}.

\begin{table*}[!t]
\centering
\fontsize{7.5}{9}\selectfont
\setlength{\tabcolsep}{2.5pt}
\renewcommand{\arraystretch}{1.12}
\begin{tabular*}{\textwidth}{@{\extracolsep{\fill}}lcccccccccccc@{}}
\toprule
& \multicolumn{4}{c}{Single-Key NIAH-1} & \multicolumn{4}{c}{Single-Key NIAH-2} & \multicolumn{3}{c}{Multi-Key MK-NIAH} & \\
\cmidrule(lr){2-5}\cmidrule(lr){6-9}\cmidrule(lr){10-12}
Model & 1K & 2K & 4K & 8K & 1K & 2K & 4K & 8K & 1K & 2K & 4K & Avg. \\
\midrule
GDN & 92.6 & 62.0 & 32.4 & 13.2 & 19.8 & 41.6 & \textbf{13.8} & 1.6 & 24.4 & 20.8 & 12.4 & 30.42 \\
\quad +PoST & 93.2 & 64.8 & 35.6 & 14.6 & 47.6 & 47.6 & 2.4 & 9.4 & 23.6 & 21.4 & 8.2 & 33.49 \\
\quad +DAMI & \textbf{95.4} & \textbf{84.0} & \textbf{53.8} & \textbf{24.2} & \textbf{89.6} & \textbf{55.6} & 13.0 & \textbf{12.8} & \textbf{25.0} & \textbf{22.0} & \textbf{12.6} & \textbf{44.36} \\
\midrule
KDA & 63.8 & 29.4 & 16.8 & 5.6 & \textbf{88.8} & 25.6 & 1.2 & 3.8 & \textbf{23.0} & 18.2 & 14.0 & 26.38 \\
\quad +PoST & \textbf{93.6} & 72.0 & 34.2 & 9.6 & 79.8 & 16.8 & 9.6 & 5.4 & 18.2 & 17.2 & 14.0 & 33.67 \\
\quad +DAMI & 90.2 & \textbf{80.6} & \textbf{56.8} & \textbf{25.4} & 79.4 & \textbf{51.0} & \textbf{26.2} & \textbf{7.6} & 21.8 & \textbf{19.4} & \textbf{15.6} & \textbf{43.09} \\
\midrule
GDN2 & 88.2 & 51.8 & 16.8 & 2.4 & \textbf{80.2} & 26.6 & 11.2 & 0.6 & 22.4 & 18.8 & 14.0 & 30.27 \\
\quad +PoST & 96.0 & 82.0 & \textbf{46.2} & 8.8 & 40.6 & 34.2 & 10.0 & 1.2 & 22.2 & 16.8 & 10.2 & 33.47 \\
\quad +DAMI & \textbf{98.0} & \textbf{84.8} & 45.0 & \textbf{12.6} & 70.2 & \textbf{71.6} & \textbf{29.0} & \textbf{6.4} & \textbf{23.8} & \textbf{21.6} & \textbf{17.8} & \textbf{43.71} \\
\midrule
Mamba-2 & 22.8 & 5.4 & 1.6 & 1.0 & 37.0 & 14.8 & 8.6 & 0.6 & \textbf{5.8} & 6.8 & 5.0 & 9.95 \\
\quad +PoST & 24.2 & 8.4 & 2.8 & 1.8 & 64.8 & \textbf{27.6} & 6.2 & 0.8 & 4.6 & 6.6 & 3.2 & 13.73 \\
\quad +DAMI & \textbf{81.6} & \textbf{30.2} & \textbf{13.4} & \textbf{5.0} & \textbf{69.6} & 26.8 & \textbf{17.4} & \textbf{3.6} & \textbf{5.8} & \textbf{8.0} & \textbf{7.2} & \textbf{24.42} \\
\midrule
RWKV-7 & 27.0 & 20.4 & 12.0 & 4.0 & \textbf{81.8} & \textbf{62.2} & \textbf{14.0} & \textbf{5.6} & 14.2 & 11.4 & \textbf{10.0} & 23.87 \\
\quad +PoST & 42.4 & 14.4 & \textbf{20.6} & \textbf{8.4} & 78.4 & 20.6 & 4.6 & 3.4 & 12.0 & 13.4 & 9.4 & 20.69 \\
\quad +DAMI & \textbf{99.8} & \textbf{46.6} & 4.4 & 0.8 & 72.2 & 37.8 & 11.2 & 3.4 & \textbf{19.2} & \textbf{15.4} & \textbf{10.0} & \textbf{29.16} \\
\bottomrule
\end{tabular*}
\caption{Answer-containment accuracy (\%) on the fixed v1 RULER-style NIAH benchmark.
Each task--length setting contains the same 500 pregenerated samples for all
models. Bold denotes the best result within each architecture; Avg. is the
unweighted mean over the 11 settings.}
\label{tab:main-results}
\end{table*}

\paragraph{Initialization schemes.}
Each architecture compares \textsc{Random}, \textsc{PoST}, and \textsc{DAMI}.
Random uses the baseline initialization. Our PoST condition applies
PoST-style spectral initialization \citep{cao2026optimal}, retaining
Random's \texttt{dt\_bias} and excluding position-adaptive scaling.
Its spectrum spans 0--8.318 for GDN-family models and Mamba-2; RWKV-7 solves
frozen scales for geometric timescales.
DAMI allocates geometric head-wise timescales from 1 to 4,096 tokens and a
low--high--low layer profile (approximately 0.25--0.75--0.25 sensitivity
for the GDN family).
Section~\ref{sec:dami} and Appendix~\ref{app:settings} detail the adaptations
and configurations.

\paragraph{Evaluation protocol.}
Primary retrieval uses a fixed RULER-style NIAH subset \citep{hsieh2024ruler}:
NIAH-1/2 at 1K, 2K, 4K, and 8K; MK-NIAH at 1K, 2K, and 4K.
All models share 500 pregenerated samples per setting and fixed decoding
settings. We report answer-containment accuracy and its unweighted mean over
the 11 settings.
General-capability evaluations use WikiText \citep{merity2017wikitext}, WinoGrande
\citep{sakaguchi2020winogrande}, ARC-Easy/Challenge \citep{clark2018arc},
OpenBookQA \citep{mihaylov2018openbookqa}, and Social IQa
\citep{sap2019socialiqa}. Six real-world retrieval tasks at 2K follow
BASED/GDN \citep{arora2024based,yang2025gdn}.
All evaluations use the same 15 trained checkpoints across five architectures.

\subsection{Main Results}
\label{sec:main-results}

Table~\ref{tab:main-results} reports retrieval performance under matched
training budgets. DAMI improves average accuracy over \textsc{Random} by
13.95, 16.71, 13.44, 14.47, and 5.29 percentage points for GDN, KDA, GDN2,
Mamba-2, and RWKV-7, respectively. The gains over the corresponding PoST-style
spectral controls are 10.87, 9.42, 10.24, 10.69, and 8.47 points.
DAMI achieves the best average within all five architectures and the best
result in 42 of 55 task--length comparisons, including ties.

At 8K, twice the training context, DAMI improves NIAH-1 for GDN, KDA, GDN2,
and Mamba-2. RWKV-7 gains concentrate on NIAH-1 and MK-NIAH at 1K/2K.

\paragraph{General capabilities and real-world retrieval.}
Language modeling and commonsense performance remain broadly comparable
(Table~\ref{tab:common-sense}); DAMI leads 17 of 30 comparisons, including ties.
For Mamba-2/RWKV-7, WinoGrande improves by 1.82/0.79 points over Random,
and Social IQa scores are best within each architecture.

\begin{table}[!t]
\centering
\fontsize{7.5}{9}\selectfont
\setlength{\tabcolsep}{1.4pt}
\renewcommand{\arraystretch}{1.12}
\begin{tabular*}{\columnwidth}{@{\extracolsep{\fill}}lcccccc@{}}
\toprule
Model & WikiText & WinoG. & ARC-E & ARC-C & OBQA & SIQA \\
& ppl$\downarrow$ & acc$\uparrow$ & acc$\uparrow$ & acc$\uparrow$ & acc$\uparrow$ & acc$\uparrow$ \\
\midrule
GDN & \textbf{44.44} & 48.93 & 43.81 & \textbf{25.00} & 31.60 & 36.64 \\
\quad +PoST & 44.83 & 48.78 & 43.90 & 24.74 & 30.20 & \textbf{37.15} \\
\quad +DAMI & 45.27 & \textbf{52.09} & \textbf{44.07} & 23.72 & \textbf{31.80} & 37.00 \\
\midrule
KDA & 42.04 & \textbf{51.30} & \textbf{44.11} & \textbf{24.32} & 30.60 & 37.56 \\
\quad +PoST & 42.42 & 51.22 & 43.52 & 24.06 & \textbf{31.40} & 37.10 \\
\quad +DAMI & \textbf{41.98} & 50.20 & 43.27 & \textbf{24.32} & 30.60 & \textbf{37.77} \\
\midrule
GDN2 & 41.13 & 50.75 & 44.44 & 23.72 & 31.20 & 37.62 \\
\quad +PoST & 41.64 & 49.96 & 44.40 & 24.06 & \textbf{31.40} & 37.82 \\
\quad +DAMI & \textbf{40.68} & \textbf{51.07} & \textbf{45.16} & \textbf{24.49} & 29.80 & \textbf{38.28} \\
\midrule
Mamba-2 & 49.47 & 49.64 & \textbf{45.54} & 23.38 & 31.20 & 37.05 \\
\quad +PoST & \textbf{49.25} & 50.36 & 42.68 & \textbf{24.40} & 30.60 & 36.54 \\
\quad +DAMI & 49.71 & \textbf{51.46} & 43.60 & 23.89 & \textbf{31.60} & \textbf{37.41} \\
\midrule
RWKV-7 & \textbf{43.34} & 50.12 & 44.07 & 23.89 & 29.60 & 36.69 \\
\quad +PoST & 43.74 & 50.51 & \textbf{44.23} & 24.06 & \textbf{31.00} & 36.39 \\
\quad +DAMI & 44.84 & \textbf{50.91} & 43.90 & \textbf{24.57} & 29.80 & \textbf{37.15} \\
\bottomrule
\end{tabular*}
\caption{Language modeling and commonsense results. WikiText reports perplexity
(lower is better); other tasks report accuracy (\%, higher is better).
Bold denotes the best result within each architecture. WinoG., ARC-E, ARC-C,
OBQA, and SIQA denote WinoGrande, ARC-Easy, ARC-Challenge, OpenBookQA, and
Social IQa, respectively.}
\label{tab:common-sense}
\end{table}

The six 2K real-world retrieval tasks also remain broadly comparable
(Table~\ref{tab:realworld-retrieval}). Mamba-2 with DAMI leads SQuAD, FDA,
and DROP, raising its average from 3.89 to 4.28; RWKV-7 averages span
5.67--6.42 across schemes.

\begin{table}[!t]
\centering
\fontsize{7.5}{9}\selectfont
\setlength{\tabcolsep}{1.8pt}
\renewcommand{\arraystretch}{1.12}
\begin{tabular*}{\columnwidth}{@{\extracolsep{\fill}}lccccccc@{}}
\toprule
Model & SWDE & SQuAD & FDA & TQA & NQ & DROP & Avg. \\
\midrule
GDN & \textbf{9.18} & 18.53 & \textbf{2.00} & 0.17 & \textbf{0.14} & 2.16 & \textbf{5.36} \\
\quad +PoST & 6.48 & 18.10 & 1.54 & 0.23 & 0.11 & \textbf{2.49} & 4.83 \\
\quad +DAMI & 7.47 & \textbf{19.81} & 1.09 & \textbf{0.24} & \textbf{0.14} & 2.25 & 5.16 \\
\midrule
KDA & \textbf{9.81} & \textbf{21.31} & 1.18 & \textbf{0.31} & \textbf{0.11} & 2.33 & \textbf{5.84} \\
\quad +PoST & 8.91 & 19.44 & \textbf{2.18} & 0.26 & 0.08 & 2.30 & 5.53 \\
\quad +DAMI & 8.46 & 20.27 & 1.27 & 0.07 & 0.03 & \textbf{2.35} & 5.41 \\
\midrule
GDN2 & \textbf{10.35} & 21.75 & 1.91 & 0.33 & \textbf{0.42} & 2.10 & 6.14 \\
\quad +PoST & 7.74 & 21.51 & \textbf{2.00} & 0.09 & 0.08 & \textbf{2.38} & 5.64 \\
\quad +DAMI & 9.99 & \textbf{23.49} & 1.81 & \textbf{0.34} & 0.14 & 2.28 & \textbf{6.34} \\
\midrule
Mamba-2 & \textbf{5.76} & 14.18 & 0.64 & \textbf{0.47} & 0.14 & 2.13 & 3.89 \\
\quad +PoST & 5.22 & 14.41 & 0.27 & 0.37 & \textbf{0.36} & 2.23 & 3.81 \\
\quad +DAMI & 5.22 & \textbf{16.96} & \textbf{0.91} & 0.11 & 0.22 & \textbf{2.28} & \textbf{4.28} \\
\midrule
RWKV-7 & \textbf{9.36} & \textbf{23.56} & 2.81 & \textbf{0.21} & \textbf{0.08} & 2.50 & \textbf{6.42} \\
\quad +PoST & 8.91 & 22.08 & \textbf{2.99} & 0.18 & 0.03 & \textbf{2.72} & 6.15 \\
\quad +DAMI & 8.37 & 20.78 & 2.45 & 0.06 & 0.03 & 2.34 & 5.67 \\
\bottomrule
\end{tabular*}
\caption{Real-world retrieval at 2K tokens (\%). TQA denotes TriviaQA; Avg. is
the unweighted mean over six tasks. Bold denotes the best result within each
architecture.}
\label{tab:realworld-retrieval}
\end{table}

\subsection{Ablation Studies}
\label{sec:ablations}

\paragraph{Need for Joint Initialization.}
Table~\ref{tab:ablation-axes} compares GDN Random/DAMI with variants restoring
one parameter to Random while retaining the other's DAMI value; references
reuse the main table. Throughout the ablations, NIAH averages lengths within
each family, Overall averages 11 settings, and bold marks column maxima.

\begin{table}[!htb]
\centering
\fontsize{7.5}{9}\selectfont
\setlength{\tabcolsep}{2pt}
\renewcommand{\arraystretch}{1.12}
\begin{tabular*}{\columnwidth}{@{\extracolsep{\fill}}lcccc@{}}
\toprule
Variant & NIAH-1 & NIAH-2 & MK-NIAH & Overall \\
& Avg.$\uparrow$ & Avg.$\uparrow$ & Avg.$\uparrow$ & Avg.$\uparrow$ \\
\midrule
Random & 50.05 & 19.20 & 19.20 & 30.42 \\
Random \texttt{A\_log} & 49.15 & 2.40 & 14.87 & 22.80 \\
Random \texttt{dt\_bias} & 42.55 & 16.90 & 17.27 & 26.33 \\
\textbf{DAMI} & \textbf{64.35} & \textbf{42.75} & \textbf{19.87} & \textbf{44.36} \\
\bottomrule
\end{tabular*}
\caption{GDN single-parameter replacement ablations (\%). The named parameter is restored to Random; the other retains its DAMI value.}
\label{tab:ablation-axes}
\end{table}

Full DAMI achieves an Overall accuracy of 44.36, compared with 30.42 for Random
and 22.80/26.33 for randomizing \texttt{A\_log}/\texttt{dt\_bias} alone.
Randomizing \texttt{A\_log} yields NIAH-2 accuracy of 2.40.
The advantage of the joint configuration is consistent with the coupling
in Equation~\eqref{eq:gate}.

\paragraph{Discussion on Time Scales.}
We evaluate the effects of timescale coverage and spacing.
Restricting \(\tau\) to 1--2048 or 1--1024 reduces Overall accuracy to 29.07 or
23.13 (Table~\ref{tab:tau-variants}). Linear spacing yields NIAH-1 accuracy
close to DAMI (62.40 versus 64.35), but NIAH-2 accuracy is 5.65.
These comparisons support the selected coverage and allocation.

\begin{table}[!htb]
\centering
\fontsize{7.5}{9}\selectfont
\setlength{\tabcolsep}{2pt}
\renewcommand{\arraystretch}{1.12}
\begin{tabular*}{\columnwidth}{@{\extracolsep{\fill}}lcccc@{}}
\toprule
Variant & NIAH-1 & NIAH-2 & MK-NIAH & Overall \\
& Avg.$\uparrow$ & Avg.$\uparrow$ & Avg.$\uparrow$ & Avg.$\uparrow$ \\
\midrule
$\tau$: 1--2048 & 47.95 & 17.30 & 19.60 & 29.07 \\
$\tau$: 1--1024 & 22.05 & 25.50 & \textbf{21.40} & 23.13 \\
Linear spacing & 62.40 & 5.65 & 15.47 & 28.96 \\
\textbf{DAMI} & \textbf{64.35} & \textbf{42.75} & 19.87 & \textbf{44.36} \\
\bottomrule
\end{tabular*}
\caption{Effects of timescale coverage and spacing on GDN retrieval (\%).}
\label{tab:tau-variants}
\end{table}

\begin{figure}[!htb]
\centering
\includegraphics[width=\columnwidth]{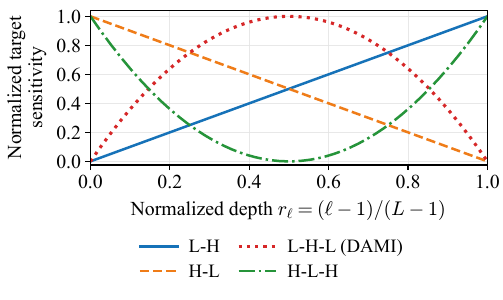}
\caption{Normalized target sensitivity profiles across layers. The horizontal
axis is normalized depth \(r_{\ell}=(\ell-1)/(L-1)\); the vertical axis is
\((\mathcal{S}^{\star}_{\ell}-\mathcal{S}_{\min})/(\mathcal{S}_{\max}-\mathcal{S}_{\min})\).
Curves indicate initialization targets.}
\label{fig:sensitivity-shapes}
\end{figure}

\paragraph{Discussion on Sensitivity.}
We compare increasing (L-H), decreasing (H-L), low--high--low (DAMI), and
high--low--high (H-L-H) sensitivity profiles
(Figure~\ref{fig:sensitivity-shapes}).

DAMI leads all three NIAH families, reaching 44.36 Overall versus
26.85--32.85 (Table~\ref{tab:sensitivity-shapes}), supporting the
representation-motivated allocation in Section~\ref{sec:dtbias-response}.

\begin{table}[!htb]
\centering
\fontsize{7.5}{9}\selectfont
\setlength{\tabcolsep}{2pt}
\renewcommand{\arraystretch}{1.12}
\begin{tabular*}{\columnwidth}{@{\extracolsep{\fill}}lcccc@{}}
\toprule
Shape & NIAH-1 & NIAH-2 & MK-NIAH & Overall \\
& Avg.$\uparrow$ & Avg.$\uparrow$ & Avg.$\uparrow$ & Avg.$\uparrow$ \\
\midrule
L-H & 54.70 & 19.55 & 19.40 & 32.29 \\
H-L & 50.55 & 9.15 & 18.87 & 26.85 \\
\textbf{L-H-L (DAMI)} & \textbf{64.35} & \textbf{42.75} & \textbf{19.87} & \textbf{44.36} \\
H-L-H & 60.05 & 16.60 & 18.27 & 32.85 \\
\bottomrule
\end{tabular*}
\caption{GDN retrieval with four layer-wise sensitivity profiles (\%).}
\label{tab:sensitivity-shapes}
\end{table}

Shifting sensitivity by \(\pm0.2\) or moving its peak to layer 4/13 yields
24.31--33.31 Overall (Table~\ref{tab:sensitivity-perturbations}). DAMI leads
Overall, while Peak L4 leads MK-NIAH, showing the effect of profile level
and peak location.

\begin{table}[!htb]
\centering
\fontsize{7.5}{9}\selectfont
\setlength{\tabcolsep}{2pt}
\renewcommand{\arraystretch}{1.12}
\begin{tabular*}{\columnwidth}{@{\extracolsep{\fill}}lcccc@{}}
\toprule
Variant & NIAH-1 & NIAH-2 & MK-NIAH & Overall \\
& Avg.$\uparrow$ & Avg.$\uparrow$ & Avg.$\uparrow$ & Avg.$\uparrow$ \\
\midrule
Shift $+0.2$ & 45.25 & 6.10 & 20.67 & 24.31 \\
Shift $-0.2$ & 57.40 & 14.75 & 23.13 & 32.55 \\
Peak L4 & 51.25 & 22.25 & \textbf{24.13} & 33.31 \\
Peak L13 & 52.60 & 20.80 & 18.60 & 31.76 \\
\textbf{DAMI} & \textbf{64.35} & \textbf{42.75} & 19.87 & \textbf{44.36} \\
\bottomrule
\end{tabular*}
\caption{GDN sensitivity-level and peak-location perturbations (\%). Shift denotes profile translation; Peak denotes the peak layer.}
\label{tab:sensitivity-perturbations}
\end{table}

\subsection{Case Study}
\label{sec:case-study}

\begin{figure}[!t]
\centering
\includegraphics[width=\columnwidth]{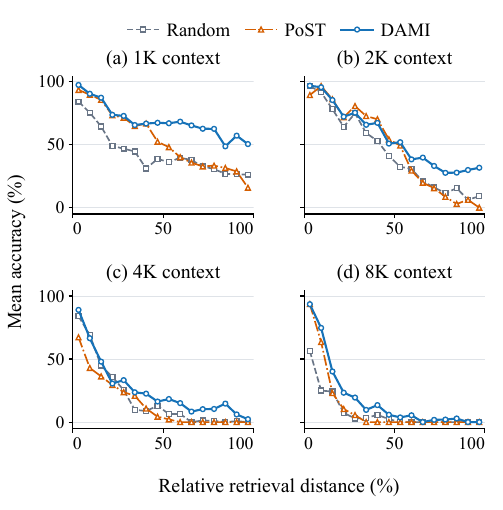}
\caption{GDN retrieval accuracy by relative distance at four context lengths.
Points show task-averaged accuracy in 16 shared bins.}
\label{fig:gdn-retrieval-distance}
\end{figure}

\paragraph{Retrieval by Distance.}
Figure~\ref{fig:gdn-retrieval-distance} compares task-averaged GDN accuracy
across relative retrieval distances at four context lengths, retaining all
500 samples per task--length setting. We normalize each target--query token
gap by the token span from the background's start to the query.
DAMI maintains higher mean accuracy in distant bins at 1K, 2K, and 4K.
Appendix~\ref{app:distance-binning} details binning and task averaging.

%% file: sections/05-conclusion.tex
\section{Conclusion}
\label{sec:conclusion}

This work presents DAMI, a dual-axis memory initialization framework motivated
by persistent correlations between initial and trained memory parameters
in controlled GDN experiments.
DAMI coordinates head-wise timescales and layer-wise input responses through
joint calibration for GDN, KDA, and GDN2, with architecture-specific extensions
to Mamba-2 and RWKV-7.
Experiments across these five recurrent architectures show improved long-context
retrieval with broadly comparable language modeling, commonsense reasoning,
and real-world retrieval performance. GDN ablations demonstrate the benefit
of jointly configuring the two memory parameters.
These findings support treating initialization as a coordinated design choice
for memory behavior, informed by the model's actual input activations.

%% file: sections/06-limitations.tex
\section*{Limitations}
This work focuses on decay-side initialization, coordinating memory
timescales and input sensitivities within existing recurrent architectures.
Our analysis and experiments characterize this design under each
architecture's native write parameterization and state-update rules.
The write side governs how incoming content is incorporated into the
recurrent state, offering a complementary perspective on memory formation
and its interaction with decay.
Future work could jointly study decay-side and write-side initialization
to clarify how their coordination balances the incorporation of new
information with the retention of useful history.

%% file: sections/appendix.tex
\section*{Appendix}
\section{Approximation Basis for Geometric Spectra}
\label{app:spectral}
For spectral coverage, Theorem~4.7 of PoST \citep{cao2026optimal} considers
approximating \(s^{-\beta}\) on \([1,T]\) with \(H\) exponentials, for
\(\beta\in(0,1)\) and \(T>1\). Let \(\Sigma_H\) be the set of such scalar
exponential sums. The paper gives a minimax exponential error rate attainable
with geometric spacing:
\begin{equation}
\begin{aligned}
\mathcal{E}_H&:=\inf_{f\in\Sigma_H}
\|s^{-\beta}-f(s)\|_{L^{\infty}[1,T]} \\
&\le C_1\exp\!\left(-\frac{\pi^2 H}{\log T+C_2}\right).
\end{aligned}
\label{eq:optimal-spectrum-error}
\end{equation}
Here \(C_1,C_2>0\) are independent of \(H\). The proof connects the problem to
rational approximation through a Laplace transform; asymptotically uniform
nodes in the log domain yield
\(\log\lambda_{h+1}-\log\lambda_h\approx\mathrm{const}\).

\section{Joint Calibration Derivation}
\label{app:calibration}
Monotonicity in the bias permits bisection in Equation~\eqref{eq:solve-dt}.
Approximating the accumulation in Equation~\eqref{eq:retention} by \(s\)
times the calibration-set mean Softplus factor and matching
Equation~\eqref{eq:timescale} gives
\begin{equation}
\begin{aligned}
\tau_{\ell,h}^{-1}&=\exp(A_{\log,\ell,h}) \\
&\quad\cdot\mathbb{E}_z\!\left[
\operatorname{softplus}(z+\mathrm{dt\_bias}_{\ell,h})\right].
\end{aligned}
\label{eq:calibration-timescale}
\end{equation}
Since \texttt{dt\_bias} changes this mean, \texttt{A\_log} must adjust jointly.
Substituting the sample mean and \(\tau_{\ell,h}=\tau_h\) yields
Equation~\eqref{eq:solve-alog}.

\section{Architecture-specific Adaptation Details}
\label{app:extensions}
For Mamba-2, we retain a geometrically spaced head-wise decay spectrum and
transfer the low--high--low layer profile by setting
\(\phi(\mathrm{dt\_bias}^{\mathrm{new}}_{\ell,h})
=m_\ell\phi(\mathrm{dt\_bias}^{\mathrm{Random}}_{\ell,h})\).
The multipliers \(m_\ell\) are resampled to 42 layers and normalized to unit
geometric mean. This static adaptation preserves all other initial tensors.

For RWKV-7's sigmoid gate, we denote the native channel bias by
\texttt{dt\_bias} and the log decay scale by \texttt{A\_log}
(\texttt{w\_lora.lora.2.bias} and \texttt{decay\_log\_scale} in code).
All three conditions use 160 frozen head--layer scale values, extending the
native constant log scale of \(-0.5\) without adding trainable parameters.
DAMI arranges the mean channel-wise sigmoid response of the static biases
in a low--high--low profile and solves the scale to match the head-wise
timescales. This target concerns the sigmoid gate response; the Softplus
derivative target in Equation~\eqref{eq:sensitivity} applies to the GDN family.
Both extensions preserve their respective state-update operators.

\section{Detailed Experimental Configurations}
\label{app:settings}

\paragraph{Model dimensions.}
All models use hidden size 512 and head dimension 64.
GDN, KDA, and GDN2 have 18 layers and six heads, with 103.79M, 101.69M,
and 112.25M parameters, respectively; each uses value expansion 2 and
width-4 short convolutions. Mamba-2 has 42 layers, 16 heads, and 105.01M
parameters, with expansion 2, state size 128, and width-4 convolution.
RWKV-7 has 20 layers, eight heads, and 101.43M parameters.
Each model is trained on 5B tokens with a maximum context of 4,096 tokens.
The selected RWKV-7 Random/PoST checkpoints use seed 42, and DAMI uses 43.

\paragraph{Initialization settings.}
The PoST condition uses a six-head spectrum spanning 0--8.318 for the
GDN family and a 16-head spectrum over the same range for Mamba-2.
RWKV-7 retains Random's channel bias and solves frozen scales for eight
geometric timescales. This condition uses spectral initialization only,
with no position-adaptive scaling.
GDN-family DAMI targets are
\(\{4096,776.05,147.03,27.86,5.28,1\}\) tokens.
Its 18-layer sensitivity profile has endpoints 0.25 and a central peak
of 0.748. KDA/GDN2 use centered channel perturbations with standard
deviation 0.03, aligning within-head means with the GDN target.
Mamba-2 resamples the layer multipliers to 42 layers and normalizes their
geometric mean to one, as in Appendix~\ref{app:extensions}.
RWKV-7 uses a 0.25--0.75--0.25 mean sigmoid-response profile and an
eight-head geometric timescale ladder from 4,096 to 1.

\paragraph{Distance-based retrieval analysis.}
\label{app:distance-binning}
Figure~\ref{fig:gdn-retrieval-distance} uses all 500 samples per task--length
setting. Each bin averages task-specific accuracies with equal weights:
NIAH-1, NIAH-2, and MK-NIAH for 1K--4K; NIAH-1 and NIAH-2 for 8K.
The 16 shared bins start at equal widths; one boundary is minimally adjusted
using sample positions alone to ensure each task contributes to every bin.
Points use bin centers.